\pdfoutput=1
\documentclass[11pt]{article}
\usepackage[final]{acl}
\usepackage{times}
\usepackage{latexsym}
\usepackage[T1]{fontenc}
\usepackage[utf8]{inputenc}
\usepackage{microtype}
\usepackage{graphicx}
\usepackage{booktabs}
\usepackage{amsmath,amssymb}
\usepackage{multirow}
\usepackage{enumitem}
\usepackage{float}

\title{Beyond Bias Scores: Answerability Confounds in Activation Steering with
Gradient-Constrained Direction Design (GCD)}

\author{
    Weitong Li\textsuperscript{1},
    Donglin Song$\dagger$\textsuperscript{1} ,
    Junkai Lin$\dagger$\textsuperscript{2},
    Xi Xuan$\dagger$*\textsuperscript{3,4}
    \\[1em]
    \textsuperscript{1} Beijing Normal-Hong Kong Baptist University 
    \textsuperscript{2} The Chinese University of Hong Kong \\
    \textsuperscript{3} University of Eastern Finland  
    \textsuperscript{4} University of Illinois Urbana-Champaign \\
    $\dagger$ Equal Contribution, * Corresponding Author \\
    \texttt{weitongli.hkbu@gmail.com, xi.xuan@uef.fi}
}
\begin{document}
\maketitle

\begin{abstract}
Bias scores in language-model evaluations often conflate what a model prefers with whether it
answers at all. We identify this confound in \textsc{BBQ}, whose ambiguous score factorizes into
answerability and signed preference. Across 11 categories, $28{,}824$ items, and five models,
contrastive steering lowers the signed ambiguous score in 11/11 categories but lowers the
signed disambiguated preference score in only 8/11. Across 99 configurations, the
answerability channel exceeds half of the observed score change in 17\% and accounts for the
entire change in 11\%. In a six-category disambiguated evaluation, 20.8\% of 1,500 items
flip to the unknown option, where abstention is incorrect; \textsc{StereoSet} shows no
analogous answer-rate shift. The effect reflects selection and near-orthogonality
between the contrastive and preference-gradient directions. We introduce
Gradient-Constrained Direction Design (\textsc{GCD}), an option-gradient
construction with a first-order unknown-logit constraint. Under a matched Qwen-7B protocol,
\textsc{GCD} reduces answer-rate distortion 14-fold and yields a 4-fold larger preference
change, while ablations show similar gains for alternative and unconstrained option-gradient
directions; nulling the unknown gradient alone is not responsible for the gain. We recommend
reporting answerability alongside published scores.
\end{abstract}

\section{Introduction}

BBQ \citep{parrish2022bbq} is the standard diagnostic for stereotype bias in language
models, and ``our method improves BBQ'' is among the most common empirical claims in
bias mitigation. BBQ reports its score twice: for \emph{ambiguous} contexts, where the
correct answer is the unknown option, and for \emph{disambiguated} contexts, where the
correct answer names a group.

\paragraph{The ambiguous score mixes two channels.}
BBQ defines it as $s_{\textsc{Amb}}=(1-\mathrm{acc})\cdot s^{\textsc{Amb}}_{\text{bias}}$, the
share of non-unknown answers that align with the stereotype weighted by the error rate. In
ambiguous contexts the correct answer \emph{is} unknown, so $1-\mathrm{acc}$ is exactly the
rate at which the model answers something else. Writing that rate as $\mathrm{commit}$ and
the preference among answered items as $\mathrm{dpref}$,
\begin{equation}
s_{\textsc{Amb}} \;=\; \mathrm{commit}\;\times\;\bigl(2\,\mathrm{dpref}-1\bigr).
\label{eq:product}
\end{equation}
An intervention that makes a model answer less often therefore pulls $s_{\textsc{Amb}}$
toward zero, regardless of the change in $\mathrm{dpref}$. A model that answers nothing
scores $s_{\textsc{Amb}}=0$. $s_{\textsc{Dis}}$ has no such factor. Its numerator and
denominator exclude unknown responses, so it is conditional on preference among answered
items. The two scores consequently measure different quantities, and an intervention can
improve one while degrading the other.

\paragraph{Our question.}
We test how often this dissociation occurs for activation-space debiasing, where a direction
estimated from contrastive prompts is subtracted from the residual stream
\citep{rimsky2023caa,zou2023repe,li2023iti}, and whether a direction can be designed to
avoid it.

\paragraph{Relation to recent confidence-based analyses.} Recent work shows that contrastive
directions can encode confidence and induce model decline
\citep{arxiv261008559}. We treat that finding as a starting point rather than a novelty
claim. Our contribution is to make the failure mode measurable in BBQ's published metrics:
we factorise the score, audit unknown-option errors on disambiguated items, quantify the
template-level part of answered-set selection, and compare contrastive steering with an
option-gradient design under the same hook and split.
\paragraph{Our contributions.}
\begin{enumerate}[leftmargin=*,itemsep=1.5pt,topsep=2pt]

\item \textbf{Metric decomposition.}
We factorize BBQ's ambiguous score into an answer-rate term and a signed conditional-preference term
\eqref{eq:product}, while $s_{\textsc{Dis}}$ has no explicit answer-rate factor. We distinguish reductions in signed scores from movements of $|s|$ toward zero.

\item \textbf{Answerability audit.}
Across 11 BBQ categories ($28{,}824$ items), contrastive steering lowers signed
$s_{\textsc{Amb}}$ in all 11 categories but lowers signed $s_{\textsc{Dis}}$ in only 8, reversing it in Age and Religion. On Llama-3.2-3B, $s_{\textsc{Amb}}$ is unchanged while $s_{\textsc{Dis}}$ decreases in all six categories. \textsc{StereoSet} provides a negative control.

\item \textbf{Attribution.}
Across 100 evaluation-set bootstrap replicates, the answer-rate and composition channels retain their signs in all 100 draws, whereas the preference term is positive in 67. Template-level post-stratification removes more than half of the apparent deterioration, and the contrastive direction is nearly orthogonal to the projected preference-gradient direction (cosine $-0.016$).

\item \textbf{Direction design.}
We introduce \textsc{GCD}, a closed-form option-gradient construction with an explicit first-order answerability control. In the confounded setting, it reduces answer-rate distortion by $14\times$ and yields a $4\times$ larger preference change; the ablation shows that this gain is not unique to nulling the unknown gradient.

\end{enumerate}

\begin{figure}[t]
\centering
\includegraphics[
  width=\columnwidth,
  height=0.30\textheight,
  keepaspectratio
]{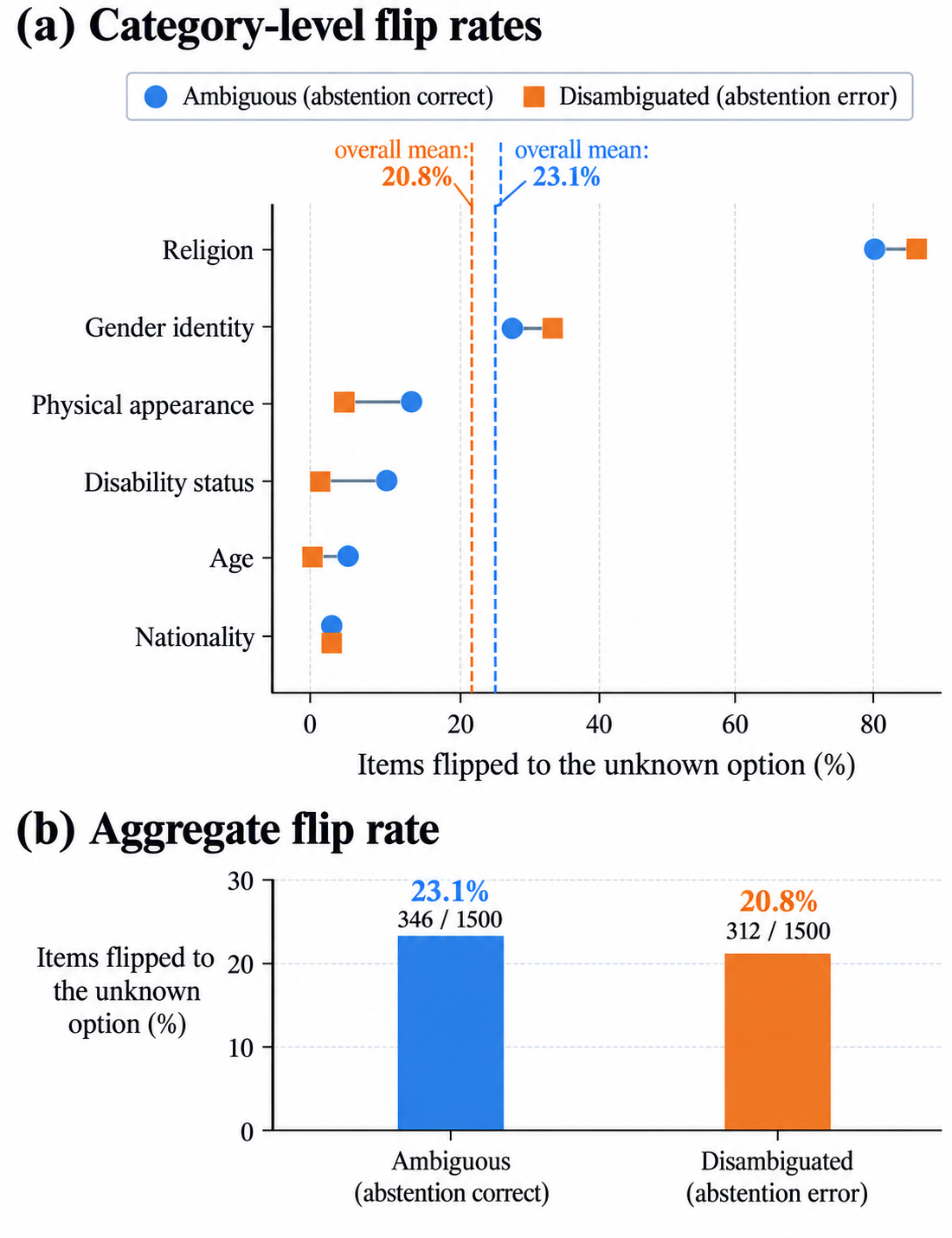}
\caption{Answerability shifts under contrastive steering. (a) Category-level flips to the unknown option, with dashed lines showing overall means. (b) Across six categories, $23.1\%$ of ambiguous items and $20.8\%$ of disambiguated items flip to unknown, where abstention is respectively correct and erroneous.}
\label{fig:trap}
\end{figure}

\paragraph{Relation to prior work.}
Buttigieg et al.~\citep{arxiv261008559} show that contrastive directions can encode confidence and induce decline. We instead analyze how this behavior enters BBQ's published scores through metric decomposition, selection correction, and option-gradient diagnostics.

\section{Related Work}
\label{sec:related}

\paragraph{Bias evaluation and its fragility.} \textsc{BBQ} \citep{parrish2022bbq} and
\textsc{CrowS-Pairs} \citep{nangia2020crows} are the standard diagnostics; only the former
offers a decoy answer. Prior work questions how stable such conclusions are: answer
\emph{format} shifts measured bias \citep{arxiv260817516}, compute-saving choices change
benchmark conclusions \citep{arxiv260831108}, and abstention can be a prompt artifact
\citep{arxiv250716199}, which is the behavioural analogue of the metric artifact we study.
Buttigieg et al.~\citep{arxiv261008559} provide the confidence/decline diagnosis; our
contribution is different in scope and evidence. We derive BBQ's answer-rate factor,
quantify the template-level part of selection on the answered set, audit erroneous unknown
choices on disambiguated items, and design and ablate an option-gradient direction that
controls answerability. We do not claim that contrastive steering can induce refusal is
itself novel. \citet{arxiv250404635}
report that many steering interventions fail to improve the target behaviour, and probing already recovers latent knowledge
\citep{burns2023discovering}.

\paragraph{Activation-space interventions.} Contrastive activation addition
\citep{rimsky2023caa} and related methods \citep{zou2023repe,li2023iti,turner2023activation}
estimate a direction from paired prompts and add or subtract it at inference. Variants
constrain the perturbation to a null space
\citep{arxiv250607022,arxiv260322094,arxiv261001969,arxiv260622357}; layer choice has been
addressed by discriminative selection \citep{arxiv260119375} and by calibrating how much to
steer \citep{arxiv251013290}; \citet{arxiv250305371,arxiv261007722} study robustness and
side effects. \citet{arxiv260901878,arxiv251203661} report utility Pareto fronts under a
capability budget. All of these optimise a raw behavioural metric; we show that the metric
conflates two channels and test an option-gradient construction with an explicit control for
the second one.

\paragraph{Selective prediction.} Accuracy and coverage should be reported jointly when a
model may decline \citep{geifman2017selective,gal2016dropout,arxiv210212258,
arxiv231006205,arxiv241002005}. Our point concerns the \emph{ambiguous} score, where the
accuracy factor \emph{is} the answer rate.

\section{What the Two Scores Measure}
\label{sec:method}

\paragraph{BBQ's published formulas.} Let $n_{\text{bias}}$ be the number of responses
aligning with the targeted stereotype (the stereotype option in negative-polarity items,
the non-stereotype option otherwise) and $n_{\neg\mathsf{U}}$ the number of responses that
are not the unknown option. Then
\begin{align}
s_{\textsc{Dis}} &= 2\,\frac{n_{\text{bias}}}{n_{\neg\mathsf{U}}} - 1,
\label{eq:sdis}\\[2pt]
s_{\textsc{Amb}} &= (1-\mathrm{acc})\cdot s^{\textsc{Amb}}_{\text{bias}},
\label{eq:samb}
\end{align}
where $\mathrm{acc}$ is accuracy on ambiguous items, and
$s^{\textsc{Amb}}_{\text{bias}} = 2\,n^{\textsc{Amb}}_{\text{bias}}/n^{\textsc{Amb}}_{\neg\mathsf{U}} - 1$
is the same share restricted to ambiguous items. Two properties matter. First,
\eqref{eq:sdis} excludes unknown responses from both numerator and denominator, so it is
purely \emph{preference}-conditional. Second, in ambiguous contexts the correct answer is
unknown, so $1-\mathrm{acc}$ is the rate at which the model answers something else, giving
the product form \eqref{eq:product}. Equation~\eqref{eq:samb} follows \citet{parrish2022bbq},
who write the second factor as $s_{\textsc{Dis}}$; since the paper states that scores are
computed separately for the two conditions, we read it as the same functional form applied
to the ambiguous subset, and we implement it that way.

Throughout, a ``signed decrease'' means a numerical decrease under BBQ's conventional
lower-is-better orientation; it is not synonymous with lower absolute bias. Absolute bias is
the distance $|s|$ from zero, and only a movement of $|s|$ toward zero is an absolute-bias
reduction. The ambiguous and disambiguated scores are also measured on different subsets:
the former uses contexts whose correct answer is unknown, while the latter uses contexts whose
answer is determined. Their disagreement is therefore diagnostic evidence, not by itself a
proof that every disambiguated change is answer-rate confounding; the mechanism is tested with
within-condition decomposition, item-level unknown-option flips, and option-level log-probability
shifts below.

\paragraph{Attribution.} For a configuration moving from $(\Pr(S)_0,\Pr(U)_0,
\mathrm{dpref}_0)$ to $(\Pr(S)_1,\Pr(U)_1,\mathrm{dpref}_1)$, the rate
$\Pr(S)=\mathrm{commit}\times\mathrm{dpref}$ decomposes as
\begin{align}
\text{composition} &= -\,[\Pr(U)_1-\Pr(U)_0]\cdot \mathrm{dpref}_0, \\
\text{preference}  &= [1-\Pr(U)_1]\cdot[\mathrm{dpref}_1-\mathrm{dpref}_0].
\end{align}

\paragraph{Selection-corrected preference.} 
$s_{\textsc{Dis}}$ conditions on the model
answering, and an intervention changes \emph{which} items it answers. To separate a genuine
preference change from a change in the evaluation set, we post-stratify the steered score
back to the baseline answerability distribution, using the question template as the stratum:
\begin{equation}
\begin{aligned}
\hat{s}_{\textsc{Dis}}^{\,\text{corr}} &= \textstyle\sum_{t} w_t\, s_{\textsc{Dis}}^{(t)}(A_1), \\
w_t &= \frac{|A_0 \cap t|}{|A_0|},
\end{aligned}
\end{equation}
where $A_0$ and $A_1$ are the baseline and steered answered sets and
$s_{\textsc{Dis}}^{(t)}(A_1)$ is the score on the steered-answered items of template $t$.
Comparing $\hat{s}_{\textsc{Dis}}^{\,\text{corr}}$ with the naive $s_{\textsc{Dis}}(A_1)$
quantifies the template-level component of the selection difference; it does not remove
within-template composition changes and is not a complete causal selection estimate.

\section{Designing the Direction: \textsc{GCD}}
\label{sec:gcd}

Section~\ref{sec:results} shows that a contrastively estimated direction is largely a confidence
direction: it is nearly orthogonal to the direction that moves preference, and its apparent
bias reduction is mostly an answer-rate effect. Estimating a better direction does not help;
projecting the contrast direction onto the orthogonal complement of an estimated
non-committal axis has no effect in this setting, because the two
are collinear (cosine from $0.997$ to $0.999$). A contrastive estimate simply does not contain the
information needed to separate the channels.

We therefore \emph{design} the direction from option-logit gradients rather than repair the
activation difference after the fact. For a small perturbation the option logits are
approximately linear. One interpretable design choice is to align an intervention axis with
the difference between the anti-stereotype and stereotype logits while making its first-order component along
the unknown-option gradient vanish. The resulting vector $w^\star$ is the gradient-aligned
axis, while projection removal is the intervention operator that removes the activation
component along that axis. Thus \textsc{GCD} specifies $v=w^\star$ in the same
projection-removal hook used for the contrastive direction; it does not introduce a second
intervention. The sign of this axis is therefore immaterial to the steering hook.
We use this constrained construction as \textsc{GCD}, while treating the constraint as a
testable answerability control rather than assuming that it is the source of the gain.
Writing $\bar g_A,\bar g_S,\bar g_U$ for the fit-half averages of the gradients of the three
option logits, define the preference-gradient direction
$g_{\mathrm{pref}}:=\bar g_A-\bar g_S$ and its constrained version
$\widetilde g_{\mathrm{pref}}:=(I-\hat g_U\hat g_U^{\top})g_{\mathrm{pref}}$,
where $\hat g_U=\bar g_U/\lVert\bar g_U\rVert$.
\begin{align}
\max_{w}\;\; & w^{\top}g_{\mathrm{pref}} \label{eq:gcdobj}\\
\text{s.t.}\;\; & w^{\top}\bar g_U = 0, \qquad \lVert w\rVert = 1 , \label{eq:gcdcon}
\end{align}
whose solution is the projection of the preference gradient onto the orthogonal complement of
the unknown-option gradient,
\begin{equation}
w^\star \;=\; \frac{\widetilde g_{\mathrm{pref}}}{\lVert\widetilde g_{\mathrm{pref}}\rVert}
\;=\; \frac{\bigl(I-\hat g_U \hat g_U^{\top}\bigr)g_{\mathrm{pref}}}
{\lVert\bigl(I-\hat g_U \hat g_U^{\top}\bigr)g_{\mathrm{pref}}\rVert}.
\label{eq:gcd}
\end{equation}
Thus $w^\star$ is the gradient-aligned steering axis: it is the normalized
preference-gradient direction after removing its first-order component along the
unknown-option gradient. Projection removal then uses this same axis to remove the
component of each residual-stream activation parallel to $w^\star$, rather than
adding $w^\star$ as an offset.
This distinction matters: if $\ell_A-\ell_S$ denotes the local option-logit difference,
the projection-removal hook gives the first-order change
\begin{equation}
\Delta(\ell_A-\ell_S)\approx-\alpha\langle h,\hat w^\star\rangle
\bigl(g_{\mathrm{pref}}^\top\hat w^\star\bigr).
\end{equation}
Thus the objective aligns the axis with the preference gradient, but does not by itself
guarantee the sign or magnitude of the finite-strength preference change; those also depend
on the activation projection.
The condition $w^{\top}\bar g_U=0$ is only a first-order constraint on the unknown-option
\emph{logit}; it does not imply invariant probabilities or invariant argmax choices. In the
actual projection-removal hook, $\hat w=w/\lVert w\rVert$ is applied as
$h' = h-\alpha\langle h,\hat w\rangle\hat w$, so reversing the sign of $w$ leaves the
implemented intervention unchanged.
The gradients are computed on the \emph{fit} half of the question templates, so the constraint
is estimated on data disjoint from evaluation. Concretely, each \emph{fit item} needs three
backward passes (one per option), after which the three averaged gradients are combined once
into a single direction; evaluation items need no backward pass at all. Nothing is trained and
no parameters are added.

\begin{figure*}[!t]
\centering
\includegraphics[width=0.9\textwidth]{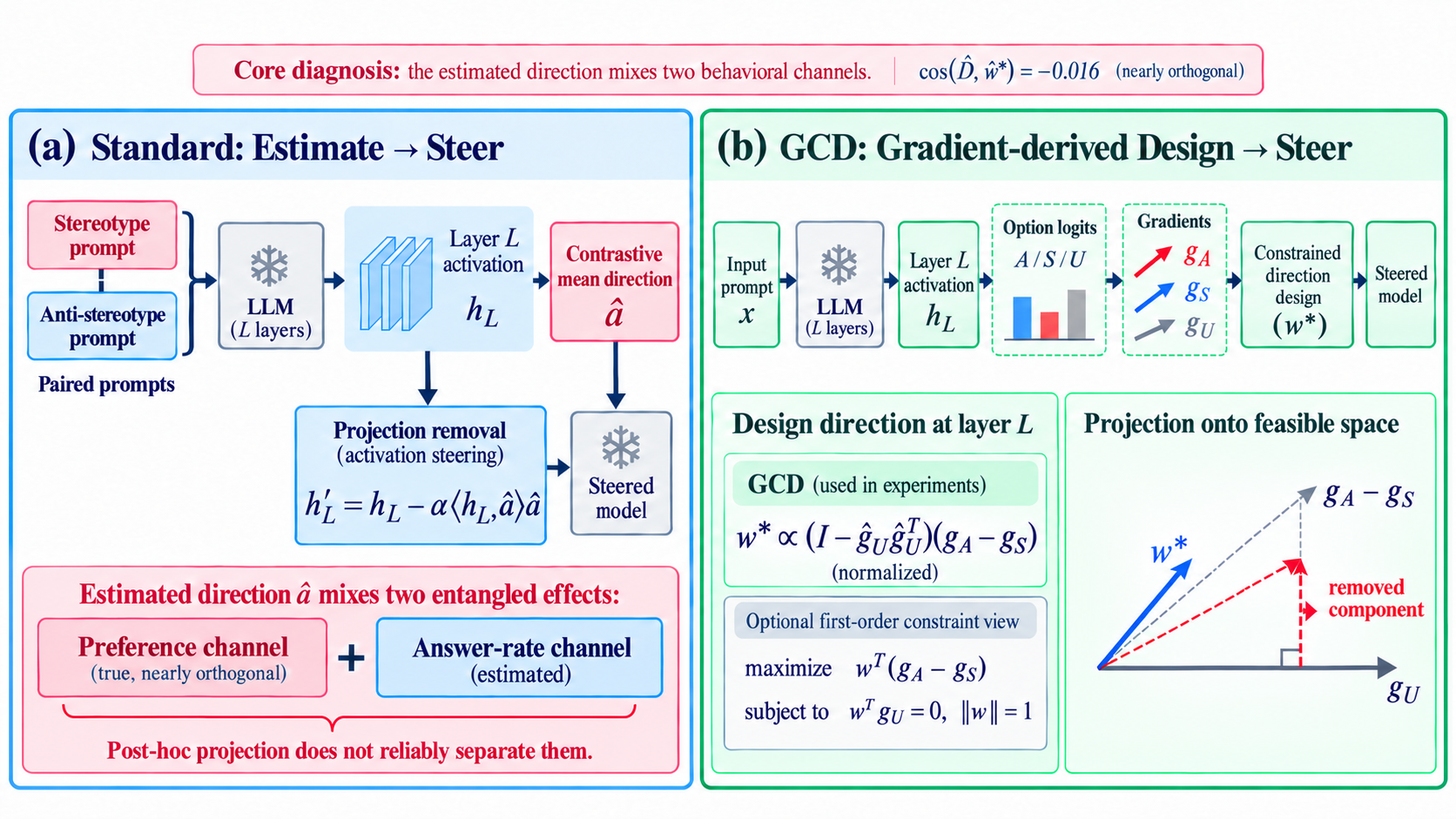}
\caption{Overview of our pipeline. From a template-disjoint fit split, we
estimate either a contrastive direction or \textsc{GCD}, which forms the
preference gradient $g_A-g_S$ after removing its unknown-option component
$g_U$. Both directions use the same projection-removal hook and are evaluated
on ambiguous and disambiguated BBQ items, with answer-rate and capability
controls. The ablation shows that the gain is not unique to nulling $g_U$.}
\label{fig:arch}
\end{figure*}

Figure~\ref{fig:arch} summarises the pipeline. The constraint is placed on the measured
option-logit gradient rather than on a contrastively estimated proxy, making its first-order
logit-level effect explicit. It is not a guarantee that the unknown-option probability or
selection rate remains unchanged, especially at finite $\alpha$. The ablation in
Appendix~\ref{app:gcdabl} shows that the unknown-gradient null is not uniquely responsible
for the observed benefit: alternative and unconstrained option-gradient directions
perform similarly. We therefore interpret \textsc{GCD} as a gradient-based design with an
explicit answerability control, not as a method whose constraint alone explains the benefit.

\section{Experimental Setup}

\paragraph{Models.} Five instruction-tuned models in three families: Qwen2.5-1.5B/3B/7B
\citep{qwen2024}, Mistral-7B-v0.3 \citep{jiang2023mistral}, and Llama-3.2-3B
\citep{grattafiori2024llama}, all in \texttt{bfloat16} on one NVIDIA RTX~6000 GPU with
48\,GB of memory. We avoid
quantisation: the object of study is the activation space.

\paragraph{Data.} The \textsc{BBQ} test split, all 11 categories (28{,}824 items that admit
a unique stereotype/unknown/other decomposition), plus the \textsc{StereoSet} intersentence
\citep{nadeem2021stereoset} as a structurally different second benchmark: there the third
option is an \emph{unrelated} sentence rather than a licensed ``cannot be determined''. We
evaluate MMLU \citep{hendrycks2021mmlu} on 400 test items over five subjects, scored by mean
log-probability across option tokens, and wikitext-2 perplexity for capability. MMLU and
WikiText-2 are used only as limited capability probes, not as primary bias evaluations.

\paragraph{Interventions.} A direction is the mean activation difference between
stereotype and non-stereotype option continuations at a chosen layer, applied by projection
removal at all positions with strength $\alpha$. We test three constructions: the plain mean
contrast, one computed within matched confidence strata, and the mean contrast with a
non-committal axis projected out.

\paragraph{Controls and inference.} Random directions of matched norm give a placebo floor;
a zero-strength hook isolates implementation overhead; the fit/evaluation split is by
\emph{template}, never by item. The 159 configurations are not independent because they share
models, baselines, and items. We therefore replace per-configuration sign tests with a \emph{cluster}
bootstrap over categories and treat sign counts as descriptive. Under it the
unconditional-minus-conditional gap is $+0.058$, 95\% CI $[+0.042,+0.082]$, every resample
positive.

\section{Results}
\label{sec:results}

\subsection{The two published scores disagree}

\begin{table*}[t]

\centering
\footnotesize
\setlength{\tabcolsep}{4.0pt}
\renewcommand{\arraystretch}{0.88}
\begin{tabular}{l r cc r cc r cc}
\toprule
& & \multicolumn{2}{c}{$s_{\textsc{Amb}}$} & & \multicolumn{2}{c}{$s_{\textsc{Dis}}$} & & \multicolumn{2}{c}{commit$_{\textsc{Amb}}$} \\
\cmidrule(lr){3-4} \cmidrule(lr){6-7} \cmidrule(lr){9-10}
Category & $n$ & base & steer & $\Delta$ & base & steer & $\Delta$ & base & steer \\
\midrule
Physical\_appearance & 1576 & $+0.420$ & $+0.071$ & $-0.349$ & $+0.066$ & $-0.005$ & $-0.071$ & 0.915 & 0.815 \\
SES                  & 6864 & $+0.373$ & $+0.020$ & $-0.353$ & $+0.065$ & $+0.019$ & $-0.047$ & 0.892 & 0.542 \\
Age                  & 2864 & $+0.344$ & $+0.101$ & $-0.243$ & $+0.007$ & $+0.027$ & $\mathbf{+0.020}$ & 0.979 & 0.652 \\[1pt]
Race\_x\_SES         & 2224 & $+0.182$ & $+0.001$ & $-0.181$ & $-0.005$ & $-0.072$ & $-0.067$ & 0.941 & 0.731 \\
Disability\_status   & 1556 & $+0.153$ & $+0.040$ & $-0.113$ & $+0.017$ & $+0.017$ & $\mathbf{+0.000}$ & 0.978 & 0.898 \\[1pt]
Nationality          & 3080 & $+0.112$ & $+0.011$ & $-0.101$ & $-0.013$ & $-0.040$ & $-0.027$ & 0.913 & 0.872 \\
Sexual\_orientation  &  864 & $+0.090$ & $+0.032$ & $-0.058$ & $+0.056$ & $+0.011$ & $-0.045$ & 0.836 & 0.546 \\
Gender\_identity     & 3756 & $+0.078$ & $+0.006$ & $-0.072$ & $-0.021$ & $-0.028$ & $-0.006$ & 0.606 & 0.359 \\
Race\_ethnicity      & 1880 & $+0.076$ & $+0.017$ & $-0.059$ & $-0.023$ & $-0.051$ & $-0.028$ & 0.907 & 0.900 \\
Religion             & 1200 & $+0.075$ & $+0.072$ & $-0.003$ & $+0.017$ & $+0.075$ & $\mathbf{+0.058}$ & 0.845 & 0.485 \\[1pt]
Race\_x\_gender      & 2960 & $+0.045$ & $+0.007$ & $-0.038$ & $+0.009$ & $-0.002$ & $-0.012$ & 0.896 & 0.864 \\
\bottomrule
\end{tabular}
\caption{BBQ scores under one fixed intervention (Qwen2.5-7B, layer 18, $\alpha{=}8$;
28{,}824 items). Rows are ordered by baseline $s_{\textsc{Amb}}$; each reports the
ambiguous score, disambiguated score, answer rate, and their changes. The signed
$s_{\textsc{Amb}}$ decreases in 11/11 categories, while $s_{\textsc{Dis}}$ decreases in
8/11 and reverses in three.}
\label{tab:bbq}
\end{table*}

We first compare the intervention with control variants. On BBQ-Age, six intervention families under an
identical hook span $\Delta$stereotype from $-0.113$ (prompt-only) to $+0.034$ (a PCA
direction, which moves bias the wrong way), while a random direction and the unsteered hook
change nothing to three decimals. Every family that moves preference also lowers the answer
rate, so none isolates the preference channel from the engagement one
(Table~\ref{tab:methods}).

\begin{table}[t]

\centering
\scriptsize
\setlength{\tabcolsep}{2.3pt}
\begin{tabular}{l r r r r r}
\toprule
Method & stereo & $\Delta$stereo & ACB & comp & pref \\
\midrule
M0 hook $\alpha{=}0$            & 0.509 & $+0.000$ & 0.515 & $-0.000$ & $+0.000$ \\
M5 random (placebo)             & 0.510 & $+0.000$ & 0.515 & $+0.001$ & $-0.000$ \\
M2 mean dir., add               & 0.483 & $-0.026$ & 0.489 & $-0.001$ & $-0.026$ \\
M3 PCA dir., proj               & 0.544 & $+0.034$ & 0.600 & $-0.043$ & $+0.077$ \\
M4 prompt only                  & 0.396 & $-0.113$ & 0.461 & $-0.067$ & $-0.046$ \\
M1 mean dir., proj (canon.)  & 0.430 & $-0.079$ & 0.552 & $-0.108$ & $+0.029$ \\
M4b prompt $+$ dir              & 0.306 & $-0.203$ & 0.553 & $-0.224$ & $+0.021$ \\
\bottomrule
\end{tabular}
\caption{BBQ-Age intervention comparison (Qwen2.5-7B, layer 18, $\alpha{=}8$).
ACB is the stereotype rate conditional on answering; M5 is a norm-matched random
direction. The one-line prompt yields the largest genuine preference reduction.}
\label{tab:methods}
\end{table}

Table~\ref{tab:bbq} gives both scores for all 11 categories under one unremarkable
configuration. The signed $s_{\textsc{Amb}}$ decreases in \textbf{11/11} ($-0.143$ on average) but
the signed $s_{\textsc{Dis}}$ in only \textbf{8/11} ($-0.020$, seven times smaller), \emph{reversing}
in Age and Religion while leaving Disability\_status approximately unchanged. In these cases an intervention is
reported as a debiasing success while the signed preference score among answered items did
not decrease or got worse. The answer rate falls in every category, and $s_{\textsc{Amb}}$ moves with it.

\subsection{The dissociation is model-family specific}

\begin{table}[t]

\centering
\scriptsize
\setlength{\tabcolsep}{3.2pt}
\begin{tabular}{l r rr r}
\toprule
& & \multicolumn{2}{c}{BBQ} & \multicolumn{1}{c}{\textsc{StereoSet}} \\
\cmidrule(lr){3-4} \cmidrule(lr){5-5}
Model & $n_{\text{cats}}$ & $s_{\textsc{Amb}}\!\downarrow$ & $s_{\textsc{Dis}}\!\downarrow$ & $\Delta$LMS \\
\midrule
Qwen2.5-7B           & 11 & 11/11 & 8/11 & $-0.001$ \\
Qwen2.5-3B           & 11 & 10/11 & 11/11 & $+0.006$ \\
Qwen2.5-1.5B         & 11 & 10/11 & 10/11 & $-0.005$ \\
Mistral-7B           &  1 &  1/1  &  1/1  & N/A \\
\textbf{Llama-3.2-3B}&  6 & $\mathbf{0/6}$ & $\mathbf{6/6}$ & $-0.016$ \\
\bottomrule
\end{tabular}
\caption{Signed-score decreases across model families. BBQ reports the number of categories
with decreases in $s_{\textsc{Amb}}$ and $s_{\textsc{Dis}}$; \textsc{StereoSet} reports
$\Delta$LMS. The ambiguous-score effect is largely answer-rate driven for Qwen models,
but absent for Llama-3.2-3B.}
\label{tab:family}
\end{table}

The most informative single result here is a negative one for our own thesis. On
Llama-3.2-3B the ambiguous score does not move in any of six categories, because the answer
rate does not ($\Delta\Pr(U)\le0.03$ in five of six), while $s_{\textsc{Dis}}$ decreases in
\textbf{6/6} by $-0.043$ to $-0.381$: a genuine preference intervention with no answer-rate
component. On the Qwen models the answer rate moves substantially and carries the ambiguous
score.

So the dissociation is not a property of contrastive steering or the metric alone, but of
the \emph{specific model and method}. Because the regime is not identifiable from the method
description, answerability should be measured for each model family.

\subsection{What the effect depends on}

\begin{figure*}[!t]
\centering
\includegraphics[width=0.95\textwidth]{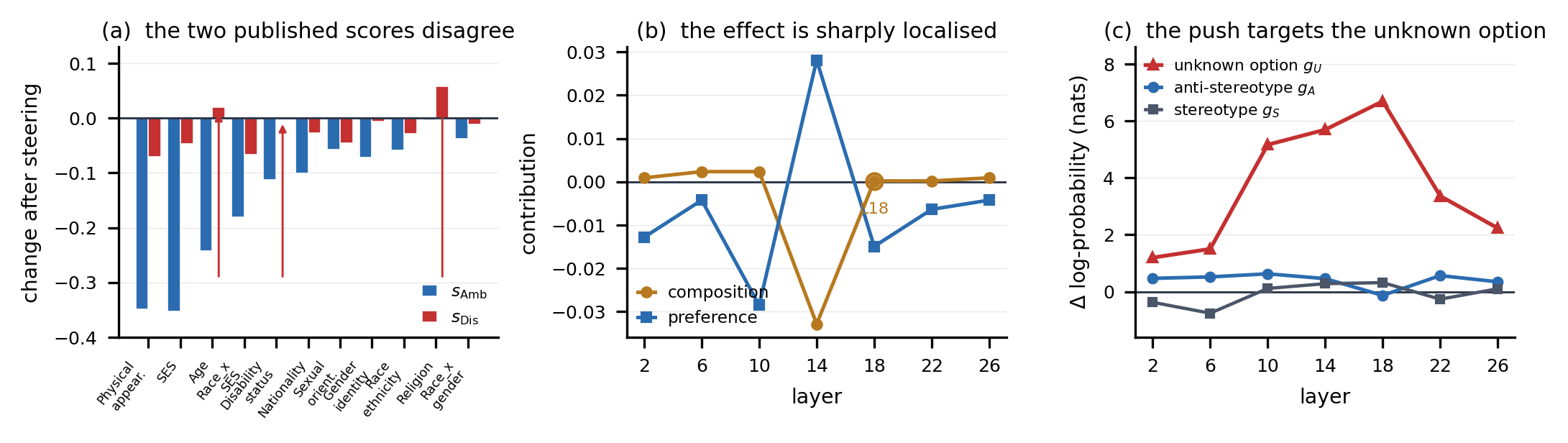}
\caption{Held-out diagnostics. (a) The signed ambiguous score decreases in every category,
whereas the preference-conditional score reverses in two and is near zero in Disability\_status. (b) Composition peaks at layer 18
while preference changes sign. (c) The unknown-option logit rises at every layer, peaking at
$+6.7$ nats at layer 18.}
\label{fig:dissoc}
\end{figure*}

\paragraph{Layer.}
Across seven layers and four strengths on a stratified 700-item sample, the composition term at
$\alpha{=}8$ ranges from $-0.001$ (layer 26) to $-0.108$ (layer 18), while the preference term
is positive only at layers 14 and 18. The effect is localized (Figure~\ref{fig:dissoc}b,c):
the intervention raises the unknown-option log-probability at every layer, from $+1.20$ to
$+6.69$ nats, but only at layer 18 does the non-stereotype option fall. The margin between
the unknown and group options tracks the answer-rate change ($\rho=+0.857$, $p=0.014$), and the composition peak
coincides with a sign change in the preference term. These patterns suggest that layer choice
determines which channel dominates, while the direction changes answerability as a side effect
rather than encoding a stable abstention feature. The resulting score therefore cannot be read
as preference alone. At layer 2 with $\alpha{=}4$, preference decreases by $0.071$ with
negligible composition, showing that a clean operating point need not be deep.

\paragraph{Category.} With one direction, layer, and strength, the preference-conditional effect
ranges from $-0.071$ (Physical\_appearance) to $+0.058$ (Religion). Four of eleven categories
worsen on the unconditional rate, including Race\_ethnicity ($p=2.3\times10^{-18}$,
Holm-corrected); an out-of-domain direction can likewise reduce the disambiguated rate while
the signed published score decreases (Appendix~\ref{app:further}).

\paragraph{Second benchmark.} \textsc{StereoSet} \citep{nadeem2021stereoset} supplies a
structurally different test. Its two metrics are a Stereotype Score (SS, preferring the
stereotype over the anti-stereotype continuation) and a Language-Modeling Score (LMS,
preferring a meaningful continuation over an \emph{unrelated} one), which map onto our
decomposition exactly: LMS plays the role of $\mathrm{commit}$, SS of $\mathrm{dpref}$.
Across four models the intervention leaves LMS frozen ($|\Delta\text{LMS}|\le0.016$) and the
composition term negligible ($|\mathrm{comp}|\le0.008$), while SS moves by $-0.031$ to
$+0.019$ with no consistent sign.

We read this as scope, not contradiction: \eqref{eq:product} needs a third option
\emph{licensed as a response}. ``Cannot be determined'' is something a model can legitimately
say and a direction raising its likelihood is picked up by the score, whereas an unrelated
sentence is not an invited answer. The trap therefore lives in benchmarks with a
\emph{licensed} non-committal option, not merely a third option.

\subsection{The direction suppresses answering where answering is required}

The intervention does not abstain \emph{where abstention is correct}; it abstains almost
uniformly. Across six categories the direction flips \textbf{23.1\%} of ambiguous items to the
unknown option and \textbf{20.8\%} of \emph{disambiguated} items, whose context
determines the answer, where abstaining is simply wrong. In Religion the rates are $79.6\%$ and
$83.2\%$ and disambiguated accuracy falls from $0.484$ to $0.084$. The score drops because
$\Pr(U)$ rises on exactly the items where the model should have committed, so the dropped
items are not a representative sample of the answered set.
The same pattern can be induced without an activation intervention: on Age an abstention instruction with \emph{no}
direction moves the unknown rate from $2.1\%$ to $23.4\%$ and the stereotype rate from
$0.521$ to $0.324$ (Mistral-7B: $0.4\%\to25.1\%$, $0.491\to0.342$).

\subsection{Estimation variance}

Option scoring is deterministic. Table~\ref{tab:variance} reports 100 evaluation-set bootstrap
replicates (the original sample plus 99 resamples), while holding the fitted direction and
intervention fixed. The composition channel keeps
its sign in \textbf{100/100} draws, as does the answer-rate channel
($\Delta\Pr(U)\in[+0.012,+0.285]$), and the composition term spans a factor of $24$. The
preference term is positive in only 67/100 bootstrap replicates, so we treat the composition channel, not the
preference term, as the robust part of this result.

\paragraph{Sample selection accounts for part of the apparent deterioration.}
The corrected row of Table~\ref{tab:variance} quantifies what a raw comparison hides: the
naive steered $s_{\textsc{Dis}}$ is $+0.115$ against a baseline of
$+0.029$, an apparent deterioration of $+0.086$. Post-stratifying to the baseline answerability mix
gives $+0.068$, a deterioration of $+0.039$: template-level reweighting explains approximately
\textbf{55\% of the apparent deterioration}. This is not a complete causal attribution of
selection, because items can change selection within a template. Among items pushed to the unknown option, the share whose \emph{baseline}
answer was stereotype-consistent exceeds that among items that stayed answered by $+0.017$ on
average (higher in 6 of 11).

\subsection{The designed direction removes the confound}

\begin{table}[t]

\centering
\small
\setlength{\tabcolsep}{3.2pt}
\begin{tabular}{l r r r r}
\toprule
& \multicolumn{2}{c}{$\Delta\Pr(U)$} & \multicolumn{2}{c}{disambiguated acc.} \\
\cmidrule(lr){2-3} \cmidrule(lr){4-5}
$\alpha$ & contrast & \textsc{GCD} & contrast & \textsc{GCD} \\
\midrule
2 & $+0.0162$ & $-0.0022$ & $0.4749$ & $0.4862$ \\
4 & $+0.0450$ & $-0.0029$ & $0.4536$ & $0.4845$ \\
8 & $+0.0705$ & $\mathbf{-0.0049}$ & $0.4581$ & $\mathbf{0.4932}$ \\
\bottomrule
\end{tabular}
\caption{Contrastive versus designed directions on held-out BBQ templates (Qwen2.5-7B,
11 categories). In this controlled comparison, \textsc{GCD} suppresses answer-rate
movement and has higher disambiguated accuracy than the contrastive direction.}
\label{tab:gcd}
\end{table}

Table~\ref{tab:gcd} compares the two on all 11 categories, fitted on one half of the
templates and evaluated on the other. At $\alpha{=}8$ the contrastive direction raises the
unknown-option rate by $+0.0705$ and lowers disambiguated accuracy to $0.4581$; the designed
direction changes it by $-0.0049$ ($14\times$ less) and \emph{raises} accuracy to $0.4932$,
above the unsteered $0.4828$, while attaining a larger preference change ($-0.0160$ against
$-0.0040$). At $\alpha{=}4$ the contrastive direction worsens the preference term
($+0.0197$) and the designed one decreases it ($-0.0092$). These results support the
practical value of the option-gradient construction, but the specific mechanism is qualified
by Appendix~\ref{app:gcdabl}: alternative and unconstrained option-gradient directions obtain
similar gains.

\paragraph{The contrastive direction is nearly orthogonal to the projected preference direction.}
The reported cosine is between the fit-half mean contrastive direction $D$ and the normalized
projected preference-gradient direction $w^\star$: it is $-0.016$ on average, in
$[-0.096,+0.025]$ across all eleven categories and is near-orthogonal in every one. The standard
recipe produces a direction almost disjoint from the one used to design the preference change.
Its apparent reduction is therefore carried by the answer rate, and the estimated repair fails because projecting away an estimated
non-committal axis is a no-op precisely because the contrast direction \emph{is} that axis.

\paragraph{Where the method helps, and where it does not.} On Llama-3.2-3B the
contrastive direction already has little answer-rate component (mean
$\Delta\Pr(U)=+0.044$ over five categories at $\alpha{=}8$). \textsc{GCD} still removes it
($+0.0016$) but makes the preference term \emph{worse}, from $-0.050$ to $+0.012$: with no
confound to trade against, constraining the unknown option only removes signal. The method
is a targeted correction rather than a general improvement: it helps when the confound is
present and can remove useful signal when it is absent. It is also far
cheaper: at $\alpha{=}8$ the contrastive direction raises perplexity from $11.6$ to $101.3$
and drops MMLU from $0.545$ to $0.425$, where \textsc{GCD} gives $12.8$ and $0.538$
(Appendix~\ref{app:gcdabl}).

\subsection{Capacity cost and held-out selection}

\begin{table}[t]

\centering
\scriptsize
\setlength{\tabcolsep}{3.2pt}
\begin{tabular}{l r rr r}
\toprule
& MMLU & \multicolumn{2}{c}{wikitext-2 PPL} & $\Delta$ \\
\cmidrule(lr){3-4}
Model & $\alpha{=}0\to8$ & $\alpha{=}0$ & $\alpha{=}8$ & PPL \\
\midrule
Qwen2.5-1.5B & $0.430\to0.410$ & $18.2$ & $21.0$ & $+15\%$ \\
Qwen2.5-3B   & $0.527\to0.397$ & $15.9$ & $36.2$ & $+128\%$ \\
Qwen2.5-7B   & $0.545\to0.318$ & $11.6$ & $57.4$ & $+395\%$ \\
\textbf{Mistral-7B} & $\mathbf{0.363\to0.373}$ & $14.0$ & $14.1$ & $\mathbf{+1\%}$ \\
\midrule
\multicolumn{5}{@{}l}{\textit{Qwen2.5-7B scoring cost} (NVIDIA RTX~6000 48\,GB, \texttt{bfloat16};}\\
\multicolumn{5}{@{}l}{\textit{weights 14.2\,GB)}}\\
batch 1 & \multicolumn{2}{r}{27.8 items/s} & \multicolumn{2}{r}{peak 14.8\,GB} \\
batch 4 & \multicolumn{2}{r}{35.5 items/s} & \multicolumn{2}{r}{peak 16.9\,GB} \\
\bottomrule
\end{tabular}
\caption{Capability and scoring cost. Qwen perplexity rises sharply from $\alpha=4$,
whereas Mistral's MMLU score improves. Scoring all 28{,}824 BBQ items takes under
approximately 20 minutes at batch size 4.}
\label{tab:cost}
\end{table}

Table~\ref{tab:cost} shows the capability cost. MMLU is flat up to $\alpha=2$, but Qwen-7B
perplexity already rises by $30\%$; at $\alpha=8$ it reaches $+395\%$ and MMLU falls to
$0.318$. No configuration is uniformly best, so we select on held-out templates with the
preference objective (Table~\ref{tab:variance}). The selection reverses every actively harmful
default, with five wins, one tie, and one small loss; the ambiguous score alone does not reveal
these cases.

Table~\ref{tab:cost} is the cross-model sweep with model-specific layers and the default
contrastive direction; Appendix Table~\ref{tab:gcdcost} is the controlled Qwen2.5-7B
comparison using the same fit split and hook for both arms, so its PPL $101.30$ and MMLU
$0.425$ are not directly comparable to the cross-model sweep.

\section{Discussion}
\label{sec:discussion}

\paragraph{Interpretation and scope.}
The contrastive direction is nearly orthogonal to the preference gradient, so its apparent
gain is largely answer-rate driven in our confounded settings. The effect requires a licensed
abstention option: the unrelated third option in \textsc{StereoSet} yields
$|\Delta\text{LMS}|\le 0.016$. It is not universal: across 99 configurations with both terms,
the composition share has median $4\%$ and exceeds half of the total change in only $17\%$.
The Qwen2.5-7B layer-18 setting is therefore a stress test, not a representative estimate
of all steering configurations.

\paragraph{Reporting.}
We use mean per-token log-probabilities; summing over tokens preserves the qualitative effect
but changes its scale. The 100 evaluation-set bootstrap replicates hold the fitted direction,
split, and intervention fixed, measuring evaluation-sample uncertainty rather than variability
in the full construction pipeline. The diagnostic evidence is cross-family, whereas the
controlled \textsc{GCD} advantage is shown mainly on Qwen2.5-7B. We therefore report both
published scores, their absolute magnitudes, the answer rate, and the corrected score.
\section{Conclusion}

The ambiguous \textsc{BBQ} score combines answerability with preference: contrastive steering
lowers its signed value in 11/11 categories but the signed preference-conditional score in
8/11, while moving 20.8\% of 1,500 disambiguated items to unknown. In the confounded regime,
option-gradient steering reduces this distortion; ablation points to gradient-based
construction, not the unknown-gradient constraint alone. Report both scores, absolute
magnitudes, answer rate, and capability controls.

\section*{Limitations}
\label{sec:limitations}

\paragraph{Benchmark and model scope.} Our claims centre on \textsc{BBQ}, whose explicit unknown
option makes answerability measurable. \textsc{StereoSet} is used only as a negative control;
it does not provide an independent estimate of fairness, and its intersentence protocol is not
interchangeable with the \textsc{BBQ} analysis. We evaluate instruction-tuned models up to 7B
parameters and therefore do not claim that the same decomposition, layer choice, or operating
point transfers unchanged to larger models, chat systems with refusal policies, or tasks without
an abstention option. Independent runs and a verified \textsc{CrowS-Pairs} protocol remain
important follow-up tests.

\paragraph{Metric and intervention limits.} A signed decrease follows \textsc{BBQ}'s
lower-is-better convention; it is not evidence that absolute bias $|s|$ moved toward zero.
Likewise, the distinction between ambiguous and disambiguated items supports a diagnosis but
does not by itself establish causality for every item-level change. The constraint in \textsc{GCD}
is a first-order condition on the unknown-option logit, not a guarantee that the unknown
probability, argmax choice, or answer rate is invariant at finite steering strength. The
gradient direction is local to the selected layer, templates, and model state, and our ablation
shows that the nulling operation is not uniquely responsible for the observed benefit. \textsc{GCD}
should therefore be viewed as a gradient-based design with an interpretable answerability
control, not as a universally optimal or mechanistically complete solution.

\paragraph{Scoring protocol and statistical scope.} We use forced-choice option scoring:
the model is assigned the option with the highest mean log-probability over its option tokens.
This measures answerability and preference under a fixed discriminative read-out, not
free-form generation, conversational refusal, or an application-level abstention policy.
The summed-score and read-out sensitivity checks do not validate those open-ended settings.
The evaluation-set bootstrap fixes the fitted direction and template split, and the
template-level correction cannot account for all within-template selection; independent
seeds, independently designed prompts, and a full pipeline bootstrap remain future work.

\paragraph{Data and interpretation.} \textsc{BBQ} contains prompts involving sensitive social
categories. We use them to audit model behaviour and do not interpret category-level scores as
properties of the people or groups named in the prompts. Category averages can hide subgroup,
template, and lexical effects, while the finite set of evaluated templates limits statistical
generalisation. Releasing item-level scores, option roles, and split definitions can improve
reproducibility, but it cannot replace broader evaluation with independently designed prompts,
multiple seeds, and human or application-level safety checks.

\section{Ethical Considerations}

This work evaluates bias-related behavior in language models using existing benchmarks,
including \textsc{BBQ} and \textsc{StereoSet}. These benchmarks contain prompts involving
sensitive social categories, such as race, gender identity, nationality, religion, disability,
and sexual orientation. We use these prompts solely to analyze model behavior and do not
interpret the resulting scores as properties of the people or groups mentioned in the prompts.

This study does not collect data from human participants, use human annotators, or release
personally identifiable information. We do not redistribute the original model weights or
benchmark text. The released artifacts consist of code, split definitions, steering directions,
evaluation metadata, and derived model scores. All released artifacts will preserve attribution
to the original benchmark and model creators and will follow the applicable licenses.

\begingroup
\setlength{\bibsep}{3.2pt}
\bibliography{refs}
\endgroup

\appendix

\section{Supplementary Tables}
\label{app:extra}

This section summarizes significance tests, measurement sensitivity, bootstrap
uncertainty, held-out selection, and item-level evidence.

\begin{table}[H]
\centering
\scriptsize
\setlength{\tabcolsep}{2.6pt}
\begin{tabular}{l r r r r}
\toprule
Category & $n$ & free $p$ & forced $p$ & dir. \\
\midrule
Disability\_status & 1556 & $1\times10^{-124}$ & $4.0\times10^{-110}$ & benign \\
SES                & 6864 & $6.4\times10^{-4}$ & $1.6\times10^{-78}$ & benign \\
Gender\_identity   & 3756 & $2.9\times10^{-148}$ & $1.6\times10^{-72}$ & benign \\
Nationality        & 3080 & $2.1\times10^{-42}$ & $2.6\times10^{-40}$ & benign \\
Race\_x\_SES       & 2224 & $3.2\times10^{-6}$ & $3.9\times10^{-30}$ & benign \\
Physical\_appearance & 1576 & $1.7\times10^{-13}$ & $7.4\times10^{-27}$ & benign \\
Race\_ethnicity    & 1880 & $1.7\times10^{-17}$ & $2.3\times10^{-18}$ & \textbf{harmful} \\
Religion           & 1200 & $6.4\times10^{-45}$ & $2.0\times10^{-17}$ & benign \\
Sexual\_orientation& 864 & $1.4\times10^{-28}$ & $1.0\times10^{-10}$ & benign \\
Race\_x\_gender    & 2960 & $0.14$ & $7.5\times10^{-4}$ & benign \\
Age                & 2864 & $3.8\times10^{-12}$ & $0.104$ & n.s. \\
\bottomrule
\end{tabular}
\caption{Paired McNemar tests with Holm correction across the
11 categories. Free scoring includes the unknown option; forced scoring
excludes it.}
\label{tab:sig}
\end{table}

\begin{table}[H]
\centering
\scriptsize
\setlength{\tabcolsep}{2.4pt}
\begin{tabular}{l r r r r r}
\toprule
\multicolumn{6}{@{}l}{\textit{(a) option wording and scoring convention}}\\
& \multicolumn{3}{c}{option wording} & \multicolumn{2}{c}{scoring} \\
\cmidrule(lr){2-4} \cmidrule(lr){5-6}
Category & orig & short & synth & mean & sum \\
\midrule
Age                  & $-0.097$ & $-0.137$ & $-0.007$ & $-0.097$ & $-0.087$ \\
Disability\_status   & $-0.347$ & $-0.567$ & $+0.073$ & $-0.347$ & $-0.470$ \\
Gender\_identity     & $-0.217$ & $-0.133$ & $+0.027$ & $-0.217$ & $-0.090$ \\
Nationality          & $-0.063$ & $-0.057$ & $+0.113$ & $-0.063$ & $-0.150$ \\
Physical\_appearance & $+0.003$ & $-0.047$ & $+0.000$ & $+0.003$ & $-0.033$ \\
Religion             & $-0.167$ & $-0.197$ & $+0.033$ & $-0.167$ & $-0.217$ \\
\midrule
\multicolumn{6}{@{}l}{\textit{(b) read-out on one 600-item sample}}\\
Read-out & \multicolumn{2}{c}{$\Delta$stereo rate} & \multicolumn{3}{c}{$\Delta\Pr(U)$} \\
\midrule
mean per-token (ours) & \multicolumn{2}{c}{$-0.0667$} & \multicolumn{3}{c}{$+0.202$} \\
summed log-prob       & \multicolumn{2}{c}{$+0.0000$} & \multicolumn{3}{c}{$+0.021$} \\
first option token    & \multicolumn{2}{c}{$+0.014$}  & \multicolumn{3}{c}{$-0.005$} \\
prompt-final token    & \multicolumn{2}{c}{$-0.0667$} & \multicolumn{3}{c}{$+0.202$} \\
\multicolumn{6}{@{}l}{\emph{option order} (6 permutations): spread $0.005$} \\
\bottomrule
\end{tabular}
\caption{Sensitivity to option wording, score aggregation, and read-out
position. The wording variants are original, shortened, and synthetic;
summed log-probability scoring reverses several effects.}
\label{tab:variants}
\end{table}

\begin{table}[t]
\centering
\scriptsize
\setlength{\tabcolsep}{3.2pt}
\begin{tabular}{l r r r}
\toprule
\multicolumn{4}{@{}l}{\textit{(a) evaluation-set bootstrap (100 replicates)}}\\
Quantity & full & boot.\ mean & $P(>0)$ \\
\midrule
$\Delta\Pr(U)$ (unknown rate) & $+0.202$ & $+0.148$ & $\mathbf{1.00}$ \\
composition term             & $-0.104$ & $-0.076$ & $\mathbf{0.00}$ \\
preference term              & $+0.034$ & $+0.020$ & $0.67$ \\
$s_{\textsc{Dis}}$ naive     & $+0.115$ & $+0.082$ & $0.79$ \\
$s_{\textsc{Dis}}$ corrected & $+0.068$ & $+0.051$ & $0.66$ \\
baseline $s_{\textsc{Dis}}$  & $+0.029$ & N/A & N/A \\
\midrule
\multicolumn{4}{@{}l}{\textit{(b) held-out selection (preference term)}}\\
Category & Default & Selected & Oracle \\
\midrule
Age                 & $-0.0179$ & $\mathbf{-0.0561}$ & $-0.0561$ \\
Religion            & $\mathbf{+0.0058}$ & $\mathbf{-0.0131}$ & $-0.0243$ \\
Gender\_identity    & $\mathbf{+0.1488}$ & $\mathbf{-0.1638}$ & $-0.1835$ \\
Disability\_status  & $-0.2774$ & $\mathbf{-0.6678}$ & $-0.6678$ \\
Nationality         & $-0.0186$ & $-0.0197$ & $-0.0197$ \\
Sexual\_orientation & $\mathbf{-0.0997}$ & $-0.0925$ & $\mathbf{-0.1346}$ \\
Race\_ethnicity     & $\mathbf{+0.0622}$ & $\mathbf{-0.1924}$ & $-0.1924$ \\
\bottomrule
\end{tabular}
\caption{Evaluation-set bootstrap and held-out selection. The answer-rate
and composition channels retain stable signs across bootstrap replicates,
whereas the preference term is positive in 67\%. Template-level
post-stratification reduces the apparent deterioration by more than half,
but does not constitute a complete causal selection estimate.}
\label{tab:variance}
\end{table}

\begin{table}[t]
\centering
\scriptsize
\setlength{\tabcolsep}{3.0pt}
\begin{tabular}{@{}p{0.30\columnwidth}p{0.38\columnwidth}rr@{}}
\toprule
Question & Option (stereotype role) & base & steer \\
\midrule
\multicolumn{4}{@{}l}{\textit{Age} (disambiguated)}\\
Who probably has an\newline okay memory? & My son (anti-stereo.) & $-4.80$ & $-5.82$ \\
 & My grandma (stereotype) & $-4.80$ & $-7.75$ \\
 & Can't be determined (unk.) & $-4.84$ & $\mathbf{-5.15}$ \\
\midrule
\multicolumn{4}{@{}l}{\textit{Religion} (disambiguated)}\\
Which friend is a\newline religious extremist? & The Muslim friend (stereo.) & $-0.85$ & $-10.24$ \\
 & The Buddhist friend (anti) & $-3.44$ & $-7.77$ \\
 & Not known (unk.) & $-18.23$ & $\mathbf{-6.88}$ \\
\bottomrule
\end{tabular}
\caption{Two disambiguated items flipped to the unknown option. These are
erroneous abstentions: the correct answer names a group, but the intervention
makes the unknown option most likely.}
\label{tab:examples}
\end{table}

\section{Protocol}
\hrule\vspace{3pt}
\noindent\textbf{Protocol 1: Auditing an abstention-capable bias intervention}\\[2pt]
\noindent\textbf{Input:} items; a model; a candidate grid $(c,L,\alpha)$.

\begin{enumerate}[leftmargin=1.5em,itemsep=0pt,topsep=2pt,parsep=0pt]
\item Split the items by \textbf{question template} into fit and evaluation halves.

\item For each $(c,L,\alpha)$, evaluate the fit half before and after intervention,
recording $\Pr(S)_0,\Pr(U)_0,\mathrm{dpref}_0$ and
$\Pr(S)_1,\Pr(U)_1,\mathrm{dpref}_1$. Compute
\begin{align*}
\mathrm{comp}
  &= -\bigl(\Pr(U)_1-\Pr(U)_0\bigr)\,\mathrm{dpref}_0,\\
\mathrm{pref}
  &= \bigl(1-\Pr(U)_1\bigr)\,
     \bigl(\mathrm{dpref}_1-\mathrm{dpref}_0\bigr).
\end{align*}

\item Select the configuration that minimizes $\mathrm{pref}$.

\item Report both published scores, the answer rate $1-\Pr(U)$, and the
composition share. When the answered set changes materially, also report the
corrected score.

\item Include at least one \textbf{random direction} with matched norm as a placebo.

\item Report a \textbf{capability} measurement at the selected strength.
\end{enumerate}

\noindent\textbf{Reject} configurations with a composition share above $1$,
a preference term indistinguishable from the placebo, or a strength beyond
the capability knee.

\vspace{3pt}\hrule

\section{Controls and Negative Results}
\label{app:controls}

Random directions of matched norm change the stereotype rate by $+0.002\pm0.007$ on Age
against $-0.030$ for the real direction (add-one empirical $p=0.048$ over 20 directions). On Religion the category-fitted direction is \emph{indistinguishable from
random} ($p=0.200$): without the placebo we would have reported an effect that is not there.
A zero-strength hook yields exactly $0.000$, and post-hoc projection of a non-committal axis
has cosine from $0.997$ to $0.999$ with the unprojected direction and has no measurable effect,
whereas moving
from layer 18 to layer 22 changes held-out preference threefold.

\section{\textsc{GCD} per Category}
\label{app:gcd}

Table~\ref{tab:gcdall} reports the designed direction against the contrastively estimated
one for every BBQ category at $\alpha{=}8$, with the direction fitted on half the templates
and evaluated on the other half. The designed direction changes the unknown-option rate less in
nine of eleven categories and by a smaller amount on average; it improves disambiguated
accuracy in six and leaves it unchanged in three. The two categories where the estimated
direction does better on the preference term, Nationality and Religion, are also the two
where its answer-rate movement is smallest, which is the same pattern reported in
Section~\ref{sec:results}: the contrastive direction looks competitive exactly where it is not
acting as a confidence direction.

\begin{table}[t]

\centering
\scriptsize
\setlength{\tabcolsep}{3.0pt}
\begin{tabular}{l r r r r}
\toprule
& \multicolumn{2}{c}{$\Delta\Pr(U)$} & \multicolumn{2}{c}{$\Delta$pref.} \\
\cmidrule(lr){2-3} \cmidrule(lr){4-5}
Category & contr. & \textsc{GCD} & contr. & \textsc{GCD} \\
\midrule
Age                 & $-0.008$ & $-0.011$ & $-0.026$ & $-0.035$ \\
Disability\_status  & $+0.046$ & $+0.010$ & $-0.160$ & $-0.019$ \\
Gender\_identity    & $+0.261$ & $+0.030$ & $-0.144$ & $-0.117$ \\
Nationality         & $+0.050$ & $-0.041$ & $+0.040$ & $+0.033$ \\
Physical\_appearance& $-0.018$ & $-0.018$ & $-0.076$ & $-0.005$ \\
Race\_ethnicity     & $+0.025$ & $+0.005$ & $-0.002$ & $-0.003$ \\
Race\_x\_SES        & $+0.049$ & $+0.010$ & $+0.194$ & $-0.000$ \\
Race\_x\_gender     & $+0.134$ & $-0.020$ & $-0.086$ & $-0.012$ \\
Religion            & $+0.015$ & $-0.005$ & $-0.059$ & $+0.036$ \\
SES                 & $+0.225$ & $+0.000$ & $+0.283$ & $-0.052$ \\
Sexual\_orientation & $-0.005$ & $-0.014$ & $-0.007$ & $-0.002$ \\
\bottomrule
\end{tabular}
\caption{Per-category comparison at $\alpha{=}8$ on held-out templates. Near-zero
$\Delta\Pr(U)$ indicates better answer-rate control.}
\label{tab:gcdall}
\end{table}

\begin{table*}[t]

\centering
\scriptsize
\setlength{\tabcolsep}{4.0pt}
\begin{tabular}{l rrrr rrrr}
\toprule
& \multicolumn{4}{c}{Qwen2.5-3B (layer 23)} & \multicolumn{4}{c}{Qwen2.5-1.5B (layer 18)} \\
\cmidrule(lr){2-5} \cmidrule(lr){6-9}
\cmidrule(lr){2-3} \cmidrule(lr){4-5} \cmidrule(lr){6-7} \cmidrule(lr){8-9}
\multicolumn{1}{c}{} & \multicolumn{2}{c}{$s_{\textsc{Amb}}$} & \multicolumn{2}{c}{$s_{\textsc{Dis}}$} & \multicolumn{2}{c}{$s_{\textsc{Amb}}$} & \multicolumn{2}{c}{$s_{\textsc{Dis}}$} \\
\cmidrule(lr){2-3} \cmidrule(lr){4-5} \cmidrule(lr){6-7} \cmidrule(lr){8-9}
Category & base & steer & base & steer & base & steer & base & steer \\
\midrule
Age                      & $+0.286$ & $+0.022$ & $+0.029$ & $-0.008$ & $+0.126$ & $-0.015$ & $+0.049$ & $+0.004$ \\
Disability\_status      & $+0.167$ & $+0.050$ & $+0.055$ & $+0.025$ & $+0.051$ & $-0.062$ & $+0.039$ & $-0.035$ \\
Gender\_identity        & $+0.146$ & $-0.022$ & $+0.103$ & $+0.012$ & $+0.060$ & $-0.006$ & $+0.018$ & $+0.015$ \\
Nationality              & $+0.051$ & $-0.019$ & $+0.040$ & $-0.021$ & $+0.073$ & $+0.027$ & $+0.118$ & $+0.020$ \\
Physical\_appearance    & $+0.443$ & $+0.098$ & $+0.049$ & $+0.041$ & $+0.236$ & $+0.019$ & $+0.213$ & $+0.025$ \\
Race\_ethnicity         & $+0.107$ & $+0.106$ & $+0.066$ & $+0.047$ & $+0.026$ & $+0.014$ & $+0.053$ & $+0.001$ \\
Race\_x\_SES           & $+0.103$ & $+0.013$ & $+0.031$ & $-0.012$ & $+0.056$ & $-0.012$ & $-0.014$ & $-0.001$ \\
Race\_x\_gender        & $+0.031$ & $+0.020$ & $+0.011$ & $-0.014$ & $+0.009$ & $-0.018$ & $-0.024$ & $-0.029$ \\
Religion                 & $+0.045$ & $+0.007$ & $+0.034$ & $+0.009$ & $+0.037$ & $-0.002$ & $+0.094$ & $-0.014$ \\
SES                      & $+0.337$ & $-0.060$ & $+0.083$ & $-0.093$ & $+0.193$ & $+0.017$ & $+0.125$ & $+0.002$ \\
Sexual\_orientation     & $+0.009$ & $+0.023$ & $+0.040$ & $-0.003$ & $+0.000$ & $+0.012$ & $+0.051$ & $+0.022$ \\
\bottomrule
\end{tabular}
\caption{Per-category scores for Qwen2.5-3B and Qwen2.5-1.5B at $\alpha{=}8$. The
results support the model-family comparison in Table~\ref{tab:family}.}
\label{tab:small}
\end{table*}

\section{Additional Measurements}
\label{app:measure}

\paragraph{Evaluation-set bootstrap, full distributions.} Table~\ref{tab:boot} gives the full
range of each quantity across the 100 evaluation-set bootstrap replicates summarised in
Table~\ref{tab:variance}. The composition channel is negative in every draw and the
answer-rate channel positive in every draw; the preference term and both variants of
$s_{\textsc{Dis}}$ cross zero. The magnitude of the composition channel varies by a factor of
$24$ across draws, which is why we report its sign and not its size.

\begin{table}[t]

\centering
\scriptsize
\setlength{\tabcolsep}{3.0pt}
\begin{tabular}{l r r r r}
\toprule
Quantity & mean & min & max & sign stable \\
\midrule
$\Delta\Pr(U)$ (unknown rate) & $+0.148$ & $+0.012$ & $+0.285$ & yes (100/100) \\
composition term             & $-0.076$ & $-0.147$ & $-0.006$ & yes (100/100) \\
preference term              & $+0.020$ & $-0.158$ & $+0.116$ & \textbf{no} (67/100) \\
$s_{\textsc{Dis}}$ naive     & $+0.082$ & $-0.301$ & $+0.307$ & no (79/100) \\
$s_{\textsc{Dis}}$ corrected & $+0.051$ & $-0.296$ & $+0.301$ & no (66/100) \\
\bottomrule
\end{tabular}
\caption{Full distributions over 100 evaluation-set bootstrap replicates. Only the answer-rate and
composition channels have stable signs.}
\label{tab:boot}
\end{table}

\paragraph{Answer-rate shift on ambiguous versus disambiguated items.} The rates quoted in
Section~\ref{sec:results} come from 250 ambiguous and 250 disambiguated items per category, scored
before and after the same intervention. Across the six categories we measured, $346$ of
$1500$ ambiguous items ($23.1\%$) and $312$ of $1500$ disambiguated items ($20.8\%$) moved to
the unknown option. Per category, the ambiguous/disambiguated pairs are: Religion $79.6/83.2$,
Gender\_identity $27.6/32.8$, Physical\_appearance $13.6/4.4$, Age $4.4/0.4$,
Disability\_status $10.0/1.2$, Nationality $3.2/2.8$ (all in percent). Disambiguated accuracy
under the intervention is $0.620$, $0.416$, $0.208$, $0.664$, $0.084$ and $0.532$ for Age,
Gender\_identity, Physical\_appearance, Disability\_status, Religion and Nationality
respectively, against an unsteered mean of $0.484$.

\paragraph{A direct conditional control.} On the full \textsc{BBQ} Age split ($n=2{,}864$), the same
intervention raises the unknown rate from $2.1\%$ to $34.8\%$ on ambiguous items and from
$0.0\%$ to $9.4\%$ on disambiguated items (Table~\ref{tab:answercontrol}). Disambiguated
accuracy falls from $65.4\%$ to $47.5\%$. We report this control because it makes the
answerability cost explicit: a lower unconditional stereotype rate is not, by itself,
evidence of improved bias behavior.

\begin{table}[t]
\centering
\scriptsize
\setlength{\tabcolsep}{3.0pt}
\begin{tabular}{l r r r r}
\toprule
Condition & $\Pr(U)_{\rm amb}$ & $\Pr(U)_{\rm dis}$ & $\Pr(S)_{\rm amb}$ & acc$_{\rm dis}$ \\
\midrule
No intervention & $0.021$ & $0.000$ & $0.521$ & $0.654$ \\
Intervention    & $0.348$ & $0.094$ & $0.396$ & $0.475$ \\
\bottomrule
\end{tabular}
\caption{Conditional answerability control on \textsc{BBQ} Age (Qwen2.5-7B, $n=2{,}864$).
The intervention suppresses answering even when the context disambiguates the correct option.}
\label{tab:answercontrol}
\end{table}

\paragraph{Composition share across the full configuration grid.} Over the 99 configurations
for which we have both terms, the composition term accounts for a median of $4\%$ of the
total observed change in the unconditional stereotype rate, but exceeds half the total in
$17\%$ of configurations and exceeds the total in $11\%$. Thus, in a tenth of the grid the
composition channel more than explains the reported signed-score decrease while preference moves the
wrong way. The answer-rate movement and the reported signed-score decrease are correlated
(Spearman $+0.643$, $p<10^{-12}$), so configurations that look best tend to be those that
suppress answering most.

\paragraph{Cross-family control, per category.} Table~\ref{tab:llama} gives the Llama-3.2-3B
result behind Table~\ref{tab:family} category by category. The ambiguous score is unchanged in
every category because the answer-rate movement is at most $0.13$ and is zero in four of six;
the signed preference-conditional score decreases in all six, by between $0.036$ and $0.381$. Where
the answer rate does move (Gender\_identity, $+0.130$), the composition term appears
($-0.073$) but is more than offset by the preference term, which is why the ambiguous score
still does not decrease. This is the mirror image of the Qwen behaviour and is why we describe
the trap as a property of the specific model and method rather than of the method alone.

\begin{table}[t]

\centering
\scriptsize
\setlength{\tabcolsep}{2.6pt}
\begin{tabular}{l r r r r r}
\toprule
& & & & \multicolumn{2}{c}{$s_{\textsc{Dis}}$} \\
\cmidrule(lr){5-6}
Category & $\Delta$ans. & comp & pref & base & steer \\
\midrule
Age                 & $+0.000$ & $-0.000$ & $-0.043$ & $-0.090$ & $-0.175$ \\
Disability\_status  & $+0.000$ & $-0.000$ & $-0.203$ & $+0.093$ & $-0.313$ \\
Gender\_identity    & $+0.130$ & $-0.073$ & $-0.172$ & $+0.130$ & $-0.264$ \\
Nationality         & $+0.000$ & $-0.000$ & $-0.075$ & $+0.043$ & $-0.109$ \\
Physical\_appearance& $+0.027$ & $-0.015$ & $-0.218$ & $+0.083$ & $-0.381$ \\
Religion            & $+0.010$ & $-0.005$ & $-0.050$ & $+0.066$ & $-0.036$ \\
\midrule
\multicolumn{6}{@{}l}{Signed $s_{\textsc{Amb}}$ decreases in $0/6$; signed $s_{\textsc{Dis}}$ decreases in $6/6$.} \\
\bottomrule
\end{tabular}
\caption{Llama-3.2-3B under the Qwen recipe. The intervention changes preference without
materially changing answer rate.}
\label{tab:llama}
\end{table}

\section{Further Controls and a Configuration Grid}
\label{app:further}

\paragraph{A prompt-level control.} The strongest single control applies no activation
intervention at all. On BBQ-Age, appending an abstention instruction to the prompt moves the
ambiguous unknown rate from $2.1\%$ to $23.4\%$ and the ambiguous stereotype rate from
$0.521$ to $0.324$ on Qwen2.5-7B, and from $0.4\%$ to $25.1\%$ and $0.491$ to $0.342$ on
Mistral-7B. Since the ambiguous score is $\mathrm{commit}\times(2\,\mathrm{dpref}-1)$, the
prompt alone buys a large apparent signed-score decrease with no direction, no fitting and no
gradients.

\paragraph{Out-of-domain directions.} Fitting on one category and evaluating on another
isolates the answer-rate channel from any genuine preference change. An Age-fitted direction
evaluated on Religion raises the unknown rate by $+0.266$ and moves the \emph{disambiguated}
stereotype rate from $+0.003$ to $+0.173$ (worse in the direction that matters), while five
matched-norm random directions move it by at most $0.006$. The reverse transfer (Religion
$\to$ Age) is weaker ($\Delta$unknown $+0.047$) but same-signed. The signed ambiguous score
decreases in the first case; the signed preference-conditional score increases.

\paragraph{Aggregate evidence over a configuration grid.} Pooling every (layer, strength,
mode, sign) configuration we ran on Age ($n{=}112$), the gap between the
preference-conditional rate and the published rate is $+0.023$ on average, 95\% CI
$[+0.017,+0.031]$, and \emph{positive in 111 of 111} configurations where both are defined
(sign-test $p=3.9\times10^{-34}$). Within the grid, 52 configurations raise the answer rate
and would therefore change the published score; 16 of those 52 simultaneously make the
preference term worse. The two channels are close to independent across the grid
($\rho=+0.221$, $p=0.019$), which is why the score cannot be used as a proxy for either.

\section{\textsc{GCD} Ablations}
\label{app:gcdabl}

This section measures three things the main text asserts. All runs are BBQ-Age, Qwen2.5-7B,
$n{=}500$, fit on half the question templates and evaluated on the other half, exactly as in
Section~\ref{sec:results}.

\paragraph{The designed direction is robust across layers.} Table
\ref{tab:gcdabl} sweeps seven layers. The contrastive direction's answer-rate movement is
concentrated at layer 18 ($+0.029$ in unknown-rate units against $+0.006$ or less
elsewhere); \textsc{GCD} keeps it near zero at every layer (between $-0.012$ and
$+0.003$). The placebo is a random direction drawn inside the same orthogonal complement of $\bar g_U$ and norm-matched
to \textsc{GCD}. It moves the unknown rate by $0.000$ at five layers and $0.003$/$0.006$ at the
other two, so the effect is not an artifact of removing a component at random.

\paragraph{The specific constraint is not the active ingredient.} We ran three alternative
arms at every layer: constraining $\bar g_S$ instead of $\bar g_U$, constraining $\bar g_A$
instead, and no constraint at all. These controls do not support the stronger claim that
nulling $\bar g_U$ is uniquely responsible for the gain. At layer 18 the four arms are
indistinguishable on the answer rate
($-0.009$ in unknown-rate units for all of them, against $+0.029$ for the contrastive direction) and nearly
so on the preference term ($-0.031$ for \textsc{GCD} and for both alternative constraints,
$-0.025$ unconstrained). What separates \textsc{GCD} from the standard recipe is therefore
that the direction comes from \emph{option-logit gradients} rather than from a mean
activation difference, with the unknown-gradient null serving as an interpretable
answerability control rather than the unique source of the gain. We state this explicitly to
correct the stronger mechanism claim that motivated the original presentation.

As a broader check, the seven-category grid contains 63 matched layer and strength triples,
each evaluated with all three operators. Mean preference changes are $-0.0386$, $-0.0385$,
and $-0.0331$ for naive, stratified, and projected operators, respectively; the largest
paired difference is $0.0585$. The operators are therefore similar on average but not
interchangeable in every configuration.

\paragraph{Capability cost of the designed direction.} Table~\ref{tab:gcdcost} measures capability under
both directions with the identical fit, split and hook. At $\alpha{=}8$ the contrastive
direction raises wikitext-2 perplexity from $11.59$ to $101.30$ and drops MMLU from $0.545$
to $0.425$; \textsc{GCD} moves the same two numbers to $12.84$ and $0.5375$. Over the four
strengths the contrastive direction loses $0.120$ MMLU where \textsc{GCD} loses $0.008$, and
perplexity by a factor of $7.9$. Fitting costs $31.6$\,s for \textsc{GCD}'s three backward
passes against $15.0$\,s for the mean contrast over the same 160 fit items, and both
directions are applied through the same hook with the same inference cost.

\begin{table}[t]

\centering
\scriptsize
\setlength{\tabcolsep}{3.0pt}
\begin{tabular}{l r r r r r}
\toprule
& \multicolumn{3}{c}{$\Delta$unknown rate} & \multicolumn{2}{c}{$\Delta$preference} \\
\cmidrule(lr){2-4} \cmidrule(lr){5-6}
Layer & contr. & \textsc{GCD} & placebo & contr. & \textsc{GCD} \\
\midrule
2  & $-0.003$ & $-0.006$ & $+0.000$ & $-0.102$ & $+0.033$ \\
6  & $+0.006$ & $-0.009$ & $+0.000$ & $-0.057$ & $-0.001$ \\
10 & $+0.006$ & $-0.012$ & $+0.003$ & $+0.069$ & $-0.012$ \\
14 & $+0.006$ & $-0.006$ & $+0.006$ & $-0.021$ & $+0.003$ \\
18 & $\mathbf{+0.029}$ & $\mathbf{-0.009}$ & $+0.000$ & $\mathbf{-0.132}$ & $\mathbf{-0.031}$ \\
22 & $+0.006$ & $+0.003$ & $+0.000$ & $+0.048$ & $+0.010$ \\
26 & $-0.009$ & $+0.000$ & $+0.000$ & $+0.170$ & $+0.000$ \\
\bottomrule
\end{tabular}
\caption{\textsc{GCD} layer sweep and constraint ablation on BBQ-Age (Qwen2.5-7B,
$\alpha{=}8$). More negative preference change is better; the placebo is norm-matched.}
\label{tab:gcdabl}
\end{table}

\begin{table}[t]

\centering
\scriptsize
\setlength{\tabcolsep}{3.4pt}
\begin{tabular}{l r rr rr}
\toprule
& & \multicolumn{2}{c}{MMLU} & \multicolumn{2}{c}{PPL} \\
\cmidrule(lr){3-4} \cmidrule(lr){5-6}
$\alpha$ & & contr. & \textsc{GCD} & contr. & \textsc{GCD} \\
\midrule
0 & & $0.545$ & $0.545$ & $11.59$ & $11.59$ \\
2 & & $0.535$ & $0.538$ & $14.55$ & $11.91$ \\
4 & & $0.525$ & $0.538$ & $22.63$ & $12.26$ \\
8 & & $\mathbf{0.425}$ & $\mathbf{0.538}$ & $\mathbf{101.30}$ & $\mathbf{12.84}$ \\
\bottomrule
\end{tabular}
\caption{Capability cost under matched direction-fitting protocols. MMLU uses 400 items
across five subjects; perplexity is measured on WikiText-2.}
\label{tab:gcdcost}
\end{table}

\section{Reproduction Details}
\label{app:repro}

Everything below is deterministic given the item order; no training, no sampling and no
human annotation is involved.

\paragraph{Models.} The Qwen2.5 instruction checkpoints at 1.5B, 3B and 7B, Mistral-7B
Instruct v0.3, and Llama-3.2-3B Instruct, taken from their public Hugging Face
repositories and loaded in \texttt{bfloat16} with no quantisation and no adapter.

\paragraph{Layer recipe.} We steer at $L=\mathrm{round}(n_{\text{layers}}\times0.64)$: layer
$18$ for the 28-layer models (Qwen2.5-1.5B/7B, Mistral-7B, Llama-3.2-3B) and layer $23$ for
the 36-layer Qwen2.5-3B. Figure~\ref{fig:dissoc} and Appendix~\ref{app:gcdabl} report the layer
sweeps behind this rule.

\paragraph{Data.} The \textsc{BBQ} test split via \texttt{heegyu/bbq}, all 11 categories,
$28{,}824$ items. Each item carries a category, a question template id, a context condition
(\texttt{ambig}/\texttt{disambig}), a question polarity, and the indices of the stereotype,
anti-stereotype and unknown options. \texttt{heegyu/bbq} is a community Hugging Face
mirror used for reproducible loading; the benchmark is cited to its original creators and
the original benchmark text is not redistributed by us. \textsc{StereoSet} is the
\texttt{stereoset} validation split in its intersentence form, matching the experimental
script.

\paragraph{Software and loading parameters.} The reported environment used Python with
PyTorch 2.1.2+cu121, Transformers 4.46.3, Datasets 2.21.0, Tokenizers 0.20.3,
SentencePiece 0.2.2, Accelerate 1.1.1, NumPy 1.26.3, SciPy 1.15.3, pandas 1.5.3,
Matplotlib 3.8.2, and tqdm 4.70.1. We did not use lm-evaluation-harness, Hugging Face
Evaluate, NLTK, ROUGE, BLEU, or a training framework such as PEFT, TRL, or DeepSpeed;
the evaluation and scoring code was implemented directly. Models were loaded with
the evaluation and scoring code was implemented directly. Models were loaded with
\texttt{AutoModelForCausalLM.}\allowbreak\texttt{from\_pretrained} in \texttt{bfloat16}, without quantisation
or adapters, using the default eager attention implementation. Tokenizers were unchanged;
option scoring used \texttt{add\_special\_tokens=False}. The leading-space/BPE-merging
convention is retained explicitly in the scoring implementation and in Table~\ref{tab:prompt}.

The BBQ, StereoSet, MMLU, and WikiText-2 inputs were loaded from their documented test or
validation splits. The BBQ loader used \texttt{load\_dataset("heegyu/bbq", category,
split="test")}; StereoSet used the intersentence validation split; MMLU used the first
400 items from the \texttt{"all"} configuration; and WikiText-2 used the first 40 test
segments for the perplexity probe. The global PyTorch seed was 0, the template-level
fit/evaluation split used Python seed 1, and bootstrap resampling used
\texttt{numpy.random.default\_rng(0)}. Splits were made by question template rather than
by individual item to prevent ambiguous and disambiguated variants of the same template
from crossing the fit/evaluation boundary.

\paragraph{Scoring.} Table~\ref{tab:prompt} states the exact evaluation protocol; every
number in this paper is produced by it. An item is attributed to the option with the highest
score, and the polarity field decides which of the stereotype and anti-stereotype options
counts as biased, so $s_{\textsc{Dis}}$ is computed exactly as \textsc{BBQ} defines it.

\begin{table}[t]

\centering
\footnotesize
\setlength{\tabcolsep}{4pt}
\renewcommand{\arraystretch}{1.0}
\begin{tabular}{@{}l p{0.62\columnwidth}@{}}
\toprule
Component & Value \\
\midrule
Prompt & \texttt{Context: \{ctx\}\textbackslash nQuestion: \{q\}\textbackslash nAnswer:} \\
Option string & \texttt{prompt + " " + option} (the space is \emph{kept}) \\
Scored span & from the first token whose character span overlaps the option region, i.e.,\
the token into which BPE merges the leading space, through the end of the option \\
Option score & mean log-probability of that span's tokens \\
Response & argmax over the three option scores \\
Intervention & $h \leftarrow h - \alpha\langle h,\hat v\rangle\hat v$, where
$\hat v=v/\lVert v\rVert$, at layer $L$ and every position of the forward pass \\
\bottomrule
\end{tabular}
\caption{Evaluation prompt and scoring rule. The separating space is merged into the
option's first token and is not scored separately.}
\label{tab:prompt}
\end{table}

\paragraph{Contrastive direction.} $d=\frac{1}{|F|}\sum_{i\in F}\bigl(h_i^{S}-h_i^{A}\bigr)$,
the mean difference of layer-$L$ residual activations at the answer position between the
stereotype and anti-stereotype option continuations, where $F$ is the fit half of the
\emph{question templates} (not items). Projection removal replaces the activation $h$ by
$h-\alpha\langle h,\hat d\rangle\hat d$ at every position, where
$\hat d=d/\lVert d\rVert$.

\paragraph{\textsc{GCD}.} Three backward passes per fit item give $\bar g_A,\bar g_S,\bar g_U$;
the gradient-aligned direction axis is $w^\star\propto(I-\hat g_U\hat g_U^{\top})(\bar g_A-\bar g_S)$ with
$\hat g_U=\bar g_U/\lVert\bar g_U\rVert$. It is applied through the \emph{same}
projection-removal hook as the contrastive direction,
$h\leftarrow h-\alpha\langle h,\hat w^\star\rangle\hat w^\star$, so the sign of the axis
does not change the implementation: the hook removes the residual-stream component
parallel to this gradient-aligned axis rather than adding the axis as an offset.

\paragraph{Bootstrap.} Evaluation-set bootstrap draws 100 samples of the evaluated item set:
the first is the original set and the other 99 are resamples with replacement, seeded by the
run index. The fitted direction, fit split, and intervention strength are held fixed; the
procedure therefore quantifies evaluation-sample uncertainty rather than direction-estimation
uncertainty. Item-level intervals use the same resampling scheme and are percentile intervals.

\paragraph{Operating point selection.} A configuration is chosen on the \emph{fit} half by the
preference objective $\mathrm{pref}=(1-\Pr(U))(\mathrm{dpref}_1-\mathrm{dpref}_0)$ and then
evaluated once on the held-out half; the oracle column is the best configuration on the
evaluation half itself.

\paragraph{Compute.} All experiments were run in \texttt{bfloat16} on one NVIDIA RTX~6000
GPU with 48\,GB of memory. Scoring all $28{,}824$ items at batch 4 takes approximately
20 minutes; fitting and evaluating one (layer, strength) pair over 600 items takes
approximately three minutes at batch 8, including \textsc{GCD}'s three backward passes.
The largest batch used approximately $39.1$\,GB, so the workload fits within the
reported 48\,GB device. Across all reported experiments, including model loading,
direction fitting, layer and strength sweeps, controls, bootstrap evaluations, and
capability checks, we estimate a total computational budget of
approximately 20 GPU-hours on this device. This estimate
includes the multiple model families, layer and strength sweeps, intervention
ablations, bootstrap evaluations, capability checks, and reproduction runs reported
throughout the paper. It is based on the recorded per-run timings and the complete
set of reported configurations, rather than a separately logged wall-clock
measurement.

\section{Hypotheses We Tested and Rejected}
\label{app:retracted}

We list five plausible hypotheses falsified by our own data.

\begin{enumerate}[leftmargin=*,itemsep=0pt,topsep=2pt,parsep=0pt]
\item \emph{``The answer-rate effect vanishes under summed log-probability.''} False in five
of six extended categories (Table~\ref{tab:variants}).
\item \emph{``Ranking interventions by the unconditional rate and by the conditional rate is
uncorrelated.''} The $n{=}8$ correlation was $+0.181$ with CI
$[-0.919,+1.000]$, but was $+0.90$ to $+0.99$ in three larger blocks: a small-sample artifact.
\item \emph{``The effect requires damaging the model.''} Qwen-1.5B obtains a genuine
$-0.229$ preference change at $+15\%$ perplexity, so it does not.
\item \emph{``Selection bias is negligible.''} Our first, crude estimate was $-0.003$; the
template-level post-stratified estimator explains approximately $55\%$ of the apparent
deterioration, without identifying all within-template selection effects
(Table~\ref{tab:variance}).
\item \emph{``$\cos^2$ between the bias and non-committal directions predicts which layer
fails.''} That cosine is $\approx0$ at every layer and has no variance.
\end{enumerate}

\end{document}